\documentclass[journal]{IEEEtran}

\usepackage{cite}
\usepackage{url}
\usepackage[hidelinks]{hyperref}

\usepackage{amsmath}
\usepackage{amssymb}
\usepackage{amsfonts}
\usepackage{bm}

\usepackage{graphicx}
\usepackage[caption=false,font=footnotesize]{subfig}

\usepackage{booktabs}
\usepackage{multirow}
\usepackage{makecell}
\usepackage{array}

\usepackage{algorithmic}

\usepackage{textcomp}
\usepackage{xcolor}

\usepackage{stfloats}

\graphicspath{{figures/}}

\begin{document}


\title{YUTELA: A Scalable Edge–Cloud AIoT System for Multi-Site Video Anomaly Detection with Scene-Specific Adaptation}



\author{
Shanle Yao,~\IEEEmembership{Student Member,~IEEE,}
Babak Rahimi Ardabili,~\IEEEmembership{Student Member,~IEEE,}
Amir Kimiyaie,
and Hamed Tabkhi,~\IEEEmembership{Member,~IEEE}%
\thanks{
The authors are with the Electrical and Computer Engineering Department, The University of North Carolina at Charlotte, Charlotte,
	NC, 28223 USA.\\
	\{syao, brahimia, htabkhiv\}@charlotte.edu\\
}
\thanks{
Corresponding author: Hamed Tabkhi.
}
}


\markboth{IEEE Internet of Things Journal}%
{Yao \MakeLowercase{\textit{et al.}}: YUTELA: A Scalable Edge–Cloud AIoT System for Multi-Site Video Anomaly Detection with Scene-Specific Adaptation}


\maketitle


\begin{abstract}

Building upon our previous Ancilia framework, this work presents \emph{YUTELA}, an edge--server--cloud AIoT system for long-term, multi-site video anomaly detection. YUTELA replaces workstation-based on-site processing with NVIDIA Jetson Thor edge devices and redesigns the multi-camera pipeline through camera batching and redistribution, shared memory, reduced inter-stage data transfer, task integration, and TensorRT-accelerated inference. It maintains a separate anomaly model per camera, retrained on a centralized GPU server once enough recent pseudo-normal data are collected, with updated weights returned to the edge. With eight cameras, TensorRT increases throughput from 14.17 to 29.94 FPS under normal density, and the optimized edge node exceeds the throughput of the workstation-based Ancilia system in seven of nine workload settings, reaching up to 2.03 times its throughput at extreme density. This gain comes at the cost of higher latency under lighter loads with four or eight streams. YUTELA was then operated continuously for 170 hours at four public sites using 31 existing CCTV cameras. The deployment shows that runtime load is driven by scene activity and input frame rate rather than by camera count, and that retraining frequency varies by about three times across sites. The 101 automatically triggered training events occupied only about 6.6\% of a single GPU, indicating that one global node can support adaptation for tens of sites. Mean end-to-end alert latency ranged from 6.8 to 15.4~s depending on site traffic and anomaly type. These results show that YUTELA enables practical multi-camera edge processing with low-cost, camera-specific adaptation across heterogeneous sites.

\end{abstract}




\section{Introduction}
\label{sec:introduction}


The rapid growth of Internet of Things (IoT)-connected cameras has created new opportunities for intelligent video surveillance systems that can perform perception, analysis, and event detection close to where video is generated ~\cite{myagmar2023survey,ardabili2023understanding}. A modern surveillance pipeline may combine object detection, pedestrian tracking, pose estimation, feature extraction, and video anomaly detection (VAD) within a single system ~\cite{REVAMP2T, ancilia}. Moving such pipelines from laboratory prototypes to long-term real-world deployment, however, remains challenging. Multiple high-bandwidth video streams must be processed continuously under limited computing and memory resources, while the visual and behavioral characteristics of each deployed camera can change substantially across locations and over time.

These challenges create two closely related requirements for practical AIoT surveillance. First, the edge-processing system must efficiently share limited computing resources across multiple cameras without introducing excessive memory use, data-transfer overhead, or processing latency ~\cite{iSENSE,wang2025edgegpu,nguyen2025octopinf}. Second, the anomaly-detection models must be able to adapt after deployment. Camera viewpoint, illumination, scene layout, pedestrian flow, and normal activity patterns can differ substantially between surveillance scenes, making it difficult for a model trained on a limited offline dataset to generalize reliably to every deployment environment ~\cite{yao2024evaluating,yang2024contextvad,zahid2025scenevad}.

Our earlier Ancilia architecture established an end-to-end AIoT framework connecting surveillance cameras, local AI processing, cloud services, and user-facing applications ~\cite{ancilia}. Its real-world evaluation further demonstrated the feasibility of operating multi-stage video analytics under practical surveillance conditions ~\cite{yao2025lab}. However, the original architecture relied on workstation-class GPU servers for local processing and treated model training largely separately from long-term deployment. These characteristics limit its use when many independent sites require compact edge hardware and camera-specific adaptation.

To address these limitations, we present \emph{YUTELA}, a multi-site edge--server--cloud AIoT system for efficient and adaptive real-world video anomaly detection. YUTELA moves on-site inference from workstation-class GPU servers to NVIDIA Jetson Thor edge devices and redesigns the processing pipeline around location-level resource sharing. Instead of assigning an isolated AI pipeline to every camera, streams from cameras at the same location are batched and redistributed across available inference pipelines. Shared CPU memory, reduced inter-stage data copying, combined low-latency processing stages, downstream rebatching, and TensorRT acceleration further reduce the memory and processing overhead of multi-camera inference.

YUTELA also extends the deployment architecture beyond fixed anomaly inference. Each surveillance camera is treated as an individual scene with its own anomaly-detection model. Recent deployment data are continuously pseudo-labeled, and pseudo-normal samples are collected for scene-specific training. Once sufficient data have been accumulated for a camera, periodic training is triggered on a centralized GPU server. The updated model weights are then returned to the corresponding edge node, allowing each camera model to adapt to its local environment without placing the training workload on the embedded device.

This separation between local inference and centralized adaptation allows multiple deployment sites to share training resources while preserving independent camera-specific models. The architecture is evaluated across four real-world sites representing an urban community college, a dining and nightlife complex, a public marketplace, and a community arts center. Each deployment processes seven to eight existing CCTV streams under different frame rates, scene activity, camera viewpoints, and operating conditions, providing a practical setting for evaluating both edge efficiency and long-term scene-specific adaptation. The main contributions of this work are summarized as follows:

\begin{itemize}

    \item We present \emph{YUTELA}, a multi-site edge--server--cloud AIoT architecture that replaces workstation-class on-site processing with embedded edge inference (Section~\ref{sec:system}).
    
    \item We introduce camera-specific adaptation triggered by accumulated pseudo-normal data volume, with centralized retraining and delivery of updated weights to the corresponding edge nodes (Section~\ref{subsec:scene_adaptation}).
    
    \item We optimize multi-camera inference through workload redistribution, shared-memory processing, stage integration, and TensorRT acceleration, achieving up to $2.03\times$ the workstation throughput under extreme density, with latency trade-offs at lighter loads (Section~\ref{subsec:edge_benchmark}).
    
    \item We evaluate YUTELA through a continuous 170-hour deployment across four public sites and 31 existing CCTV cameras, quantifying runtime performance, retraining activity, and end-to-end alert latency (Sections~\ref{subsec:realworld_deployment} and~\ref{subsec:alert_latency}).

\end{itemize}

The remainder of this paper is organized as follows. Section~\ref{sec:related_work} reviews related work in AIoT video analytics, edge intelligence, and adaptive video anomaly detection. Section~\ref{sec:system} presents the YUTELA architecture, edge-processing pipeline, and scene-specific adaptation workflow. Section~\ref{sec:experiments} describes the experimental setup and multi-site real-world evaluation, followed by the discussion and conclusion.
\section{Related Work}
\label{sec:related_work}

\subsubsection{\textbf{AIoT-Based Intelligent Video Surveillance}}
\label{subsec:related_aiot}
Real-world intelligent video surveillance systems employ optical flow\cite{11588449} \cite{SU2023110872} \cite{Alghamdi_2025_ICCV}, object detection, tracking, re-identification \cite{reidsurvey}\cite{Khan2021} , pose estimation\cite{9707520}\cite{alphapose}, action recognition\cite{SnchezCaballero2022}\cite{RW-GCN}, and anomaly detection \cite{Sreenu2019}\cite{Kim2021} algorithms to monitor activity and understand events. With recent advances in edge computing and real-time artificial intelligence, there has been a transition from running these algorithms fully on cloud to local systems closer to the network of camera sources that continuously record and transmit high volumes of visual data. Remix\cite{remix} developed a workflow for adaptive partitioning of input images that selects from a set of different neural networks on an edge device based on a latency budget. Gemel\cite{285111}, instead, merges configurations of multiple similar models on the same edge device to avoid expensive memory delays, decrease memory footprint, and improve accuracy by reducing dropped frames. DCSB\cite{dcsb} proposes a collaborative hybrid approach where edge object detection carries the bulk of the framework, while a discriminator decides if difficult cases should be sent to the big model on the cloud. RegionFilter\cite{CHEN2024110624} also implements a collaborative edge-cloud algorithm for extracting sub-segments and selecting RoIs through feedback from the cloud. While these work can be extended to handle multiple applications, they are mostly optimized and benchmarked for specific tasks rather than general context-aware surveillance.


\begin{figure*}[]
        \centering
               \includegraphics[width=0.9\linewidth, trim= 18 12 15 15,clip]{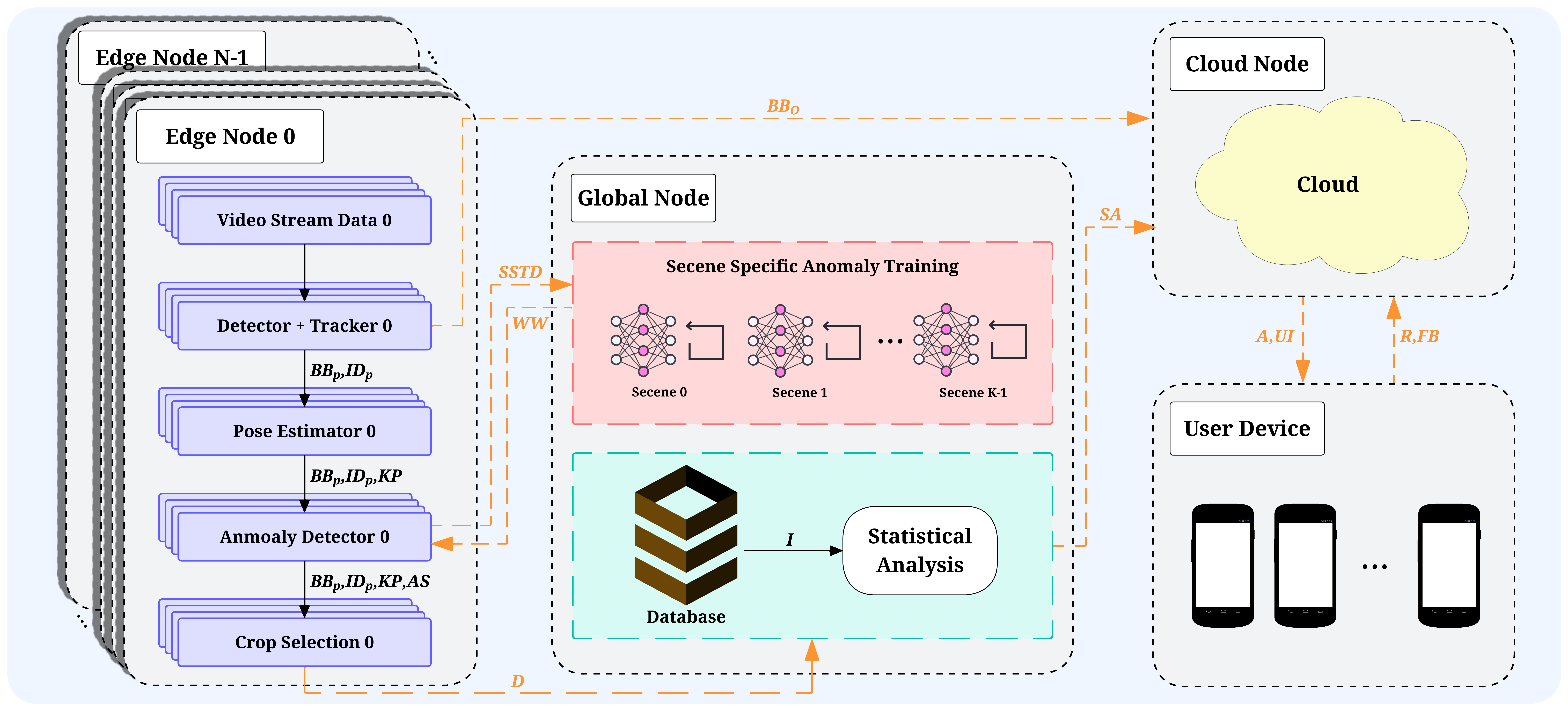}
               \caption{YUTELA system architecture and data flow. $N$ distributed edge nodes perform real-time video analysis and interact with a global node for scene-specific anomaly adaptation and multi-site statistical analysis, including behavior anomaly information, while anomaly information is transferred to the cloud for alerting and user interaction. Multiple deployment sites may share the global and cloud infrastructure, although one group is shown for clarity. $BB_P$, $ID_P$, $KP$, $AS$, $BB_O$, $D$, $SSTD$, $WW$, $I$, $SA$, $A$, $UI$, $R$, and $FB$ denote pedestrian bounding boxes, identities, keypoints, anomaly scores, object-anomaly bounding boxes, batched data, scene-specific trainable data, updated weights, database information, statistical analysis, alerts, user-interface information, requests, and feedback, respectively.}     
                \label{fig:Concept}
\end{figure*}

\subsubsection{\textbf{Video Anomaly Detection}}
\label{subsec:related_vad}

Video anomaly detection aims to detect events that deviate from normative behavior in an environment. Existing approaches adopt different representations, such as raw appearance\cite{9710957}, optical flow\cite{9710416}, scene graphs\cite{Doshi_2023_WACV}, human skeletal poses\cite{sparta}, or pretrained spatiotemporal features\cite{Zhou2023}\cite{Li2022}. These systems are trained either in an unsupervised, weakly-supervised\cite{9710957}, or fully-supervised fashion. While these methods continue to deliver promising results and useful pretrained models leveraging diverse benchmark datasets, long-term real-world deployment and benchmarks are still limited and show degraded performance. \cite{Pathirannahalage2024}\cite{yao2024evaluating}\cite{yang2024contextvad}.
Normative behavior depends heavily on the surveillance environment, with different camera viewpoints, scene layouts, pedestrian traffic, illumination conditions, and operational patterns substantially altering the data distribution between deployment sites. \cite{yao2025lab} \cite{Doshi_2022_WACV} notably noticed degraded performance of popular methods under heavy, unpredictable workloads with novel views, while \cite{JEON2024124391}\cite{Leroux_2022_WACV} highlight the impact of adverse weather and environmental conditions on real-world applications. 


\subsubsection{\textbf{Domain Adaptation and periodic training for VAD}}
\label{subsec:related_continual}

Domain shift is a central problem for practical video anomaly detection. Models trained on one dataset or surveillance site may not retain the same performance after deployment to a new environment due to factors discussed above. Hence, Continual Learning \cite{Doshi_2022_WACV} \cite{9150686} \cite{LIU2026132943} has been proposed as a solution to adapt models to changing benchmarks or environments, using fixed or adaptive criteria to select prominent novel data to learn from. Despite their decent performance, adaptation is performed in an offline fashion without systems-level implementation in mind. Additionally, retraining a new model from scratch for every camera or deployment site is expensive and difficult to maintain at scale. 

Periodic training emerges as an attractive solution by making surveillance workflows update models as new data is collected over time, while trying to preserve previously learned knowledge\cite{276952}\cite{RECL}.
However, integrating this paradigm into a distributed surveillance system poses several additional system-level questions on the choice of edge or cloud training(or both), the transfer of novel data, distributing updated weights, and the real-time inference operation as models evolve.


\subsubsection{\textbf{Summary and Research Gap}}
\label{subsec:related_gap}

Existing research has advanced edge video analytics, anomaly detection, and model adaptation, but fewer studies evaluate their integration within continuously operating, multi-site surveillance systems. Building on our Ancilia framework and its field evaluation~\cite{ancilia,yao2025lab}, YUTELA combines resource-efficient embedded inference, shared multi-camera processing, centralized camera-specific periodic training, and cloud-based alert delivery. The focus is on implementing and evaluating this complete workflow across heterogeneous real-world sites, examining computational performance, continuous operation, retraining activity, and end-to-end alert latency.

\section{YUTELA System Architecture}
\label{sec:system}

Building upon the end-to-end surveillance framework established in our previous Ancilia system~\cite{ancilia,yao2025lab}, YUTELA extends the architecture toward resource-efficient multi-site deployment and long-term adaptation to scene-dependent domain shift~\cite{huvad,yao2024evaluating}. While Ancilia demonstrated real-time video analytics using workstation-class edge servers, YUTELA redesigns the system around embedded edge computing and centralized scene-specific model adaptation. As illustrated in Fig.~\ref{fig:Concept}, the system consists of four interacting components: distributed edge nodes, a global node, a cloud service, and user devices. Video analysis is performed locally at each edge node, while selected deployment data are transferred to the global node for statistical analysis and periodic scene-specific anomaly training. Updated model weights are subsequently returned to the corresponding edge node for the specific scene, enabling the anomaly detector to adapt to local scene characteristics without interrupting continuous video processing. Object-level anomaly information is delivered through the cloud directly for alert generation, while the cloud also supports user interaction, requests, and feedback.

\subsection{Edge Node: Real-Time Video Processing}
\label{subsec:edge_node}

The edge node serves as the primary real-time inference component of YUTELA and performs video pre-processing, pedestrian and object detection, tracking, human pose estimation, and anomaly inference locally. Similar to Ancilia~\cite{ancilia}, these algorithms are organized as a sequential processing pipeline in which the output of one stage is consumed by subsequent stages. However, YUTELA further separates the handling of object-level and behavior-level anomalies according to their response-time and analysis requirements.

As illustrated in Fig.~\ref{fig:Concept}, detected object anomalies are forwarded directly from the edge node to the cloud. Object-level events generally require rapid notification; therefore, bypassing additional global processing reduces the communication and processing delay before the event becomes available to the user-facing application. Behavior anomalies follow a different path. The edge node accumulates scene-related feature information and the corresponding anomaly scores into batched data, denoted as \(D\), which are periodically transmitted to the global node. Rather than immediately treating every local anomaly score as a confirmed event, the global node performs additional statistical analysis over the collected data to reduce scene-dependent false positives. The resulting statistical analysis, denoted as \(SA\), is subsequently associated with the corresponding behavior anomaly information and forwarded to the cloud for event storage, notification, and user interaction. This separation allows YUTELA to maintain low response latency for explicit object anomalies while applying additional cross-sample analysis to behavior anomalies where false positive suppression is more important.

\begin{table}[htp]
  \footnotesize
  \centering

  \caption{Comparison of high-accuracy reference models and the algorithmic cores of YUTELA.}

  \resizebox{\columnwidth}{!}{
  \begin{tabular}{@{}c|c|c|c@{}}
    \toprule
    \toprule
    \textbf{Task} &
    \textbf{Method} &
    \textbf{Performance} &
    \textbf{Dataset} \\
    \midrule

    \multicolumn{4}{c}{\textbf{High-Accuracy Reference Algorithms}} \\
    \midrule

    Object Detection &
    Co-DETR \cite{zong2023codetr} &
    66.0 (AP) &
    COCO \cite{COCO} \\

    Tracking &
    HyperSSM \cite{song2026hyperssm} &
    80.0 (MOTA) &
    MOT20 \cite{MOT20} \\

    Pose Estimation &
    ViTPose \cite{vitpose} &
    81.1 (AP) &
    COCO \cite{COCO} \\

    Anomaly Detection &
    SPARTA \cite{sparta} &
    64.9 (AUC-ROC) &
    HuVAD\cite{huvad} \\

    \midrule
    \multicolumn{4}{c}{\textbf{Ancilia Algorithmic Core \cite{ancilia}}} \\
    \midrule

    Object Detection &
    YOLOv5 \cite{YOLOv5} &
    49.0 (mAP) &
    COCO \cite{COCO} \\

    Tracking &
    ByteTrack \cite{bytetrack} &
    77.8 (MOTA) &
    MOT20 \cite{MOT20} \\

    Pose Estimation &
    HRNet \cite{hrnet} &
    75.1 (AP) &
    COCO \cite{COCO} \\

    Anomaly Detection &
    -- &
    -- &
    -- \\

    \midrule
    \multicolumn{4}{c}{\textbf{YUTELA Algorithmic Core}} \\
    \midrule

    Object Detection &
    YOLO26 \cite{yolo26} &
    57.5 (mAP) &
    COCO \cite{COCO} \\

    Tracking &
    ByteTrack \cite{bytetrack} &
    77.8 (MOTA) &
    MOT20 \cite{MOT20} \\

    Pose Estimation &
    AlphaPose \cite{alphapose} &
    73.3 (AP) &
    COCO \cite{COCO} \\

    Anomaly Detection &
    SPARTA \cite{sparta} &
    64.9 (AUC-ROC) &
    HuVAD \cite{huvad} \\

    \bottomrule
    \bottomrule
  \end{tabular}}

  \label{tab:algorithm_core}
\end{table}

The algorithmic core of YUTELA also evolves from that of Ancilia to better support edge device deployment and behavior anomaly analysis. As summarized in Table~\ref{tab:algorithm_core}, the processing pipeline consists of object detection, pedestrian tracking, pose estimation, and anomaly detection. Because the outputs of these algorithms are propagated through the pipeline, errors introduced by an upstream model can affect subsequent processing stages. The comparison with high-accuracy reference algorithms therefore provides an indication of the accuracy tradeoff introduced by selecting models that are more practical for continuous real-time execution on edge hardware.

Compared with Ancilia, YUTELA updates the object detection and pose estimation components while retaining ByteTrack~\cite{bytetrack} for pedestrian tracking. YOLOv5~\cite{YOLOv5} and HRNet~\cite{hrnet} are replaced by YOLO26~\cite{yolo26} and AlphaPose~\cite{alphapose}, respectively, to better support deployment on the target edge platform. Rather than maximizing standalone benchmark accuracy, these models are selected to balance perception performance with the throughput, latency, and resource constraints of continuous multi-camera processing. YUTELA also introduces video anomaly detection as a major downstream task through SPARTA~\cite{sparta}, which operates on features generated by the preceding perception stages. The resulting anomaly scores and associated features form the behavior anomaly data \(D\), which is sent to the global node for further statistical analysis and scene-specific adaptation. Further details on memory-efficient and high-throughput edge inference are presented in Section~\ref{subsec:scene_adaptation}.

\subsection{Global Node: Data Management and Analysis}
\label{subsec:global_node}

The global node provides two centralized functions for edge nodes deployed across different locations: statistical data analysis and scene-specific model adaptation. Unlike the edge node, it does not participate directly in latency-sensitive video inference. Instead, it receives selected batched data and anomaly-related information from the edge nodes to support longer-term analysis and continual adaptation.

First, the global database collects batched data \(D\) from all connected edge nodes. These data include anomaly features and contextual information such as deployment location, camera ID, object ID, human count, bounding-box information, and object movement. Historical records are combined with recently received data to characterize scene activity, movement patterns, and anomaly score distributions. The resulting statistical analysis \(SA\) provides additional context for evaluating behavior anomalies and reducing false positive events before the corresponding analysis is forwarded to the cloud.

Second, the global node communicates with the anomaly detection component of each edge node to support scene-specific adaptation under domain shift. Prior studies have demonstrated the importance of scene-dependent adaptation for real-world anomaly detection~\cite{huvad,yao2024evaluating,yao2026offline}. Based on newly collected data and scene statistics, scene-specific trainable data \(SSTD\) are constructed for the corresponding deployment environment. When the adaptation criteria are satisfied, the global node performs periodic anomaly model training and generates updated model weights \(WW\), which are returned to the associated edge node. Further details on the periodic scene-specific training and domain adaptation process are presented in Section~\ref{subsec:scene_adaptation}.

\subsection{Cloud and User Interaction}
\label{subsec:cloud_user}

The cloud node provides the communication interface between the distributed surveillance infrastructure and authorized end users. Object-level anomalies are forwarded directly from the edge node to the cloud to minimize alert latency. Behavior anomaly information is delivered together with the corresponding statistical analysis \(SA\) generated by the global node. The cloud converts these events into alerts and user-interface information \((A,UI)\), which are then delivered to user devices for remote monitoring and event review.

Communication is bidirectional. User requests and feedback \((R,FB)\) are returned through the cloud to support system interaction, event review, and subsequent analysis. By separating the user-facing cloud service from the global node responsible for statistical analysis and periodic model adaptation, YUTELA keeps alert delivery and user interaction independent of the more computationally intensive training and adaptation processes.

\subsection{Edge Processing \& Scene Specific Anomaly Adaptation}
\label{subsec:scene_adaptation}

\begin{figure*}[]
    \centering
    \includegraphics[width=1\linewidth, trim=18 12 15 15,clip]{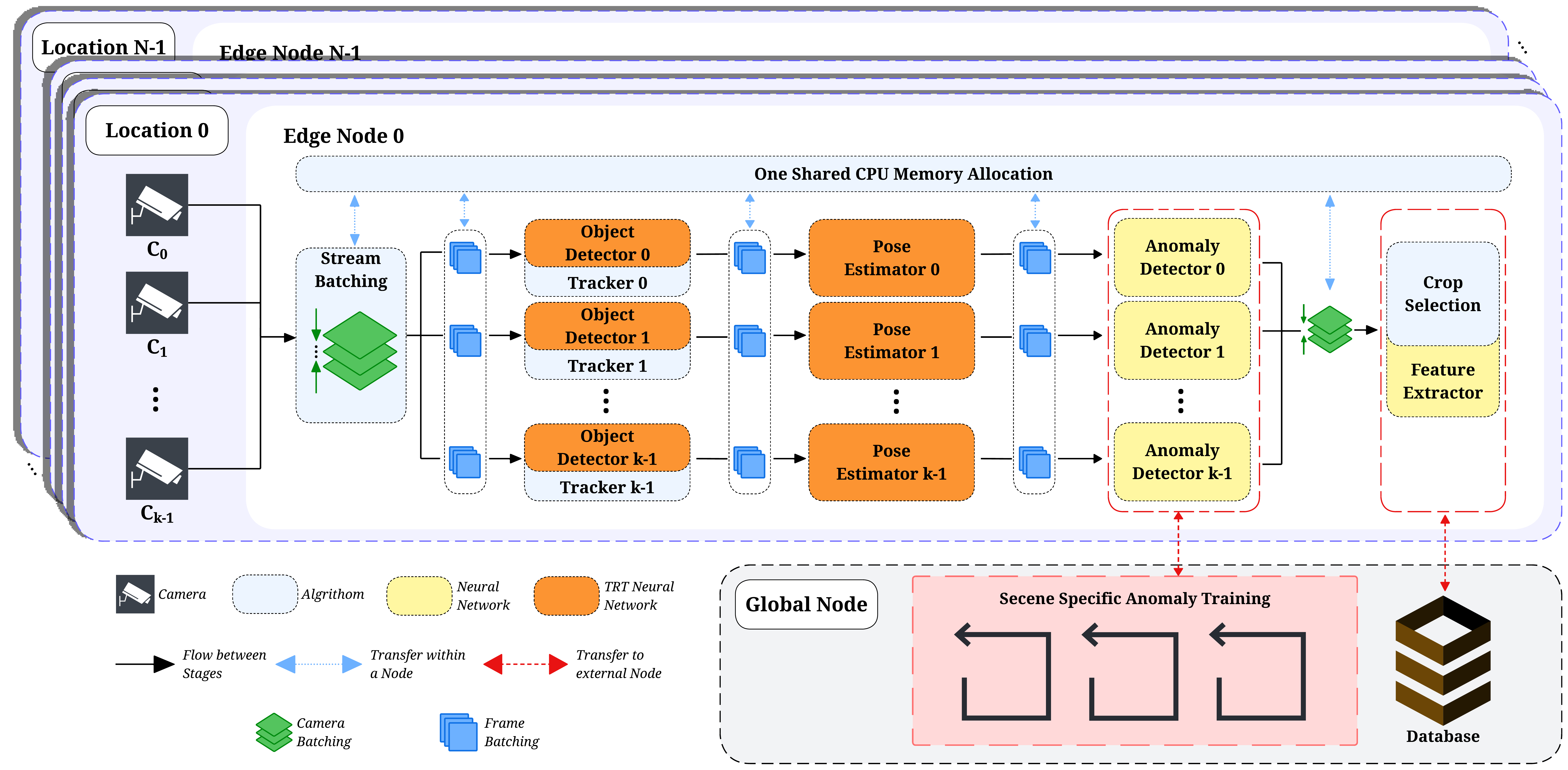}
    \caption{YUTELA edge processing and scene-specific adaptation architecture. Batched multi-camera streams share edge memory and inference resources, while anomaly features are collected at the global node for scene-specific model retraining.}
    \label{fig:pipeline}
\end{figure*}

As illustrated in Fig.~\ref{fig:pipeline}, each deployment location in YUTELA is associated with a single edge node responsible for processing all connected CCTV streams. Efficient resource sharing and workload scheduling have been explored in previous edge video analytics systems~\cite{REVAMP2T,wang2025edgegpu,nguyen2025octopinf}. Unlike a conventional design in which each camera is assigned to an isolated end-to-end AI pipeline, YUTELA first performs camera batching and redistributes incoming streams across the available processing pipelines. This design decouples the number of active pipelines from the number of cameras. Under resource-constrained configurations, multiple camera streams can therefore share the same AI pipeline, reducing duplicated model instances and improving overall hardware utilization.

After stream redistribution, YUTELA further reduces the processing overhead introduced by intermediate data movement between pipeline stages. CPU-side intermediate data are maintained within a shared memory allocation accessible across the processing stages, rather than repeatedly creating and copying redundant data between stages. This design reduces unnecessary memory allocation and CPU--GPU data transfer overhead, thereby improving end-to-end inference efficiency. The pipeline is also reorganized according to the computational characteristics of individual tasks. Lightweight operations that depend directly on upstream results are combined where possible; for example, pedestrian tracking is coupled with the object detection stage rather than being maintained as an independent processing module. Detection, pose estimation, and anomaly inference remain distributed across parallel pipelines, while their outputs are rebatched after anomaly detection. The merged data are then processed by shared downstream stages, including crop selection and feature extraction, since these operations do not require dedicated instances for each camera or inference pipeline. Selected features and associated metadata are subsequently transferred to the global node database for further analysis.

Overall, YUTELA shifts the processing architecture from a camera-to-dedicated-pipeline design toward a location-level shared-resource architecture, in which multiple camera streams are distributed across the available edge processing resources. Beyond these system-level optimizations, YUTELA further extends the anomaly detection stage with scene-specific adaptation under changing deployment conditions.

For anomaly detection, YUTELA further addresses the domain-shift problem that arises during long-term real-world deployment. Unlike benchmark datasets, which typically contain a limited number of environments, deployed surveillance systems may experience substantial variations in camera viewpoint, illumination, background activity, crowd density, and scene dynamics~\cite{yao2024evaluating,yang2024contextvad}. Consequently, a single anomaly detection model trained on a fixed offline dataset may not generalize consistently across different camera scenes~\cite{huvad,zahid2025scenevad,yao2026offline}. YUTELA therefore treats each camera as an individual domain and maintains a dedicated anomaly-detection model for each scene.

As illustrated in Fig.~\ref{fig:pipeline}, data generated by each local anomaly detector are continuously collected and pseudo-labeled following the data preparation strategy introduced in our previous work~\cite{yao2026offline,yao2025alfred}. Although incoming samples are assigned pseudo-labels, only samples identified as normal are retained for model adaptation, since the anomaly detector is trained in a self-supervised manner using normal behavior~\cite{sparta}. The collected samples remain associated with their corresponding camera, ensuring that the training data reflect the characteristics of a specific deployment scene rather than being mixed across different domains.

For each camera, scene-specific retraining is triggered after approximately one million frames containing valid human-pose observations have been accumulated, following the data-volume analysis in~\cite{yao2026offline}. Only recently collected deployment data are used to construct the scene-specific training set, allowing the model to adapt to the current characteristics of the environment. Once the required data volume is reached, the selected samples are transferred to the global node, where periodic scene-specific anomaly training is performed. The updated model weights are then returned to the corresponding edge node and replace the previous anomaly model for that camera. This closed-loop process enables each deployed anomaly detector to progressively adapt to its local scene and mitigate performance degradation caused by domain shift.
\section{Multi-Site Real-World Deployment}
\label{sec:deployment}

To evaluate YUTELA beyond controlled benchmark environments, the complete system was deployed at four real-world surveillance sites: a university campus, a local market, a downtown restaurant, and an art center. These locations represent substantially different surveillance domains in terms of camera viewpoint, scene layout, pedestrian behavior, illumination, background activity, and operating environment. Such diversity provides a practical testbed for evaluating both the efficiency of the proposed edge-processing architecture and the ability of scene-specific anomaly adaptation to address domain shift across deployments.

\begin{figure*}[t]
    \centering
    \includegraphics[width=\linewidth, trim=18 12 15 15, clip]{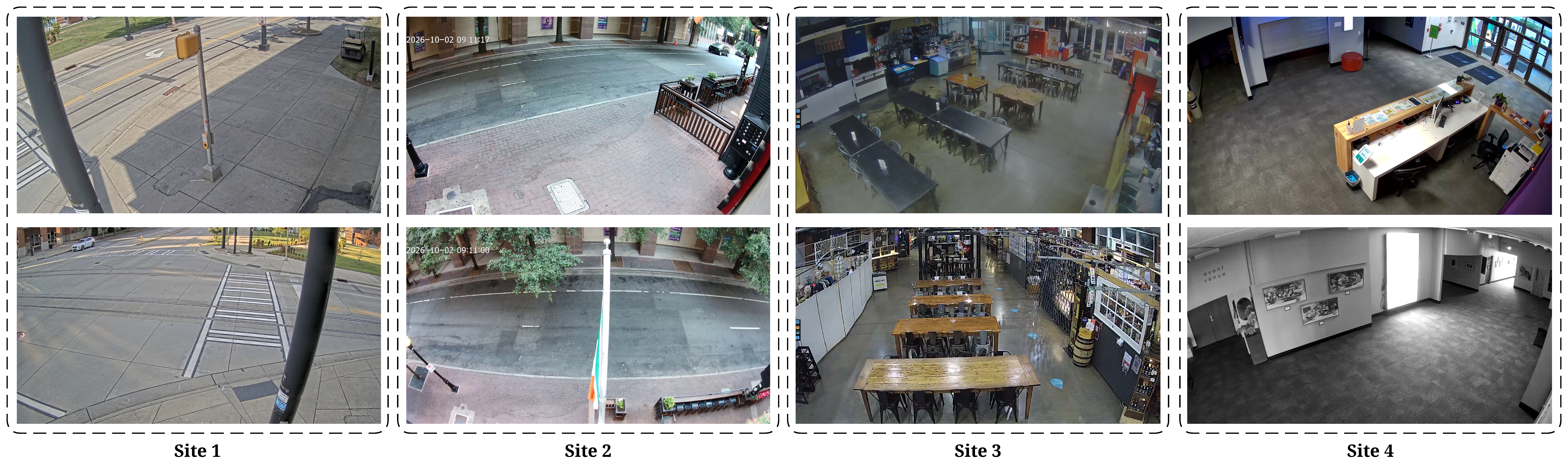}
    \caption{Representative camera views from the four real-world YUTELA deployment sites, including a university campus, local market, downtown restaurant, and art center. Two camera views are shown for each site to illustrate differences in scene layout, viewpoint, and deployment environment.}
    \label{fig:cam}
\end{figure*}

\subsection{Deployment Sites}
\label{subsec:deployment_sites}

All deployments use existing network-connected CCTV cameras already installed at the corresponding locations. YUTELA therefore operates under the camera placement, viewing angle, image quality, illumination, and scene conditions of practical surveillance systems rather than cameras installed specifically for the experiment. As shown in Fig.~\ref{fig:cam}, the four sites differ substantially in scene layout and visual characteristics. Each site provides at least seven concurrent camera streams to a local edge node, while each camera is treated as an independent scene with its own anomaly-detection model and scene specific adaptation process.

The four testbeds were selected to represent different types of public and community spaces within the same urban area. Together, they cover education, commerce, recreation, and cultural activities and provide substantially different crowd patterns, operating schedules, lighting conditions, and camera viewpoints. These differences create a useful real world setting for evaluating both multi-camera edge processing and scene specific anomaly adaptation.

\textbf{Site A -- Urban College:}
The campus environment includes outdoor walkways, road crossings, and building entrances. Pedestrian activity follows regular daytime and academic schedules, while most cameras observe relatively open scenes with moderate traffic. The deployment uses eight cameras operating at FHD and 30 FPS.

\textbf{Site B -- Dining and Nightlife Complex:}
This site contains restaurants and other entertainment spaces within a high traffic downtown environment. Camera views include entrances, pedestrian areas, and shared public spaces. Activity is generally denser than at Site A and can increase significantly during evening and late night periods. The deployment uses seven cameras operating at FHD and 30 FPS.

\textbf{Site C -- Public Marketplace:}
The marketplace contains retail, dining, and common gathering areas with frequent changes in crowd density throughout the day. Activity is often concentrated around meal periods and weekends, producing irregular and occasionally dense pedestrian flow. The deployment uses eight cameras operating at HD and 15 FPS.

\textbf{Site D -- Community Arts Center:}
The arts center includes exhibition, event, and public gathering spaces. Compared with the marketplace and nightlife site, normal activity is generally lower but can change rapidly during scheduled events. The deployment uses eight cameras operating at FHD and 20 FPS.

The selected sites also represent community-facing public environments where scalable and locally processed video analytics may provide practical value without requiring dedicated high end computing infrastructure. From a video anomaly detection perspective, these deployments provide distinct camera domains with different background structures, crowd densities, lighting conditions, activity patterns, and operating schedules. The definition of normal behavior therefore varies across both sites and individual cameras. This multi-site setting provides a realistic environment for evaluating the scene specific adaptation mechanism described in Section~\ref{subsec:scene_adaptation}.

\subsection{Edge and Global Computing Platforms}
\label{subsec:deployment_arch}

Each deployment location is equipped with an NVIDIA Jetson Thor that serves as the local YUTELA edge node. All camera streams from the corresponding site are transmitted to this device, where camera batching, perception, tracking, pose estimation, anomaly detection, and downstream feature processing are performed using the shared edge-processing architecture described in Section~\ref{subsec:scene_adaptation}. Processing multiple cameras on a single edge platform also allows the experiments to evaluate the memory and inference efficiency of YUTELA under realistic multi-camera workloads.

A centralized workstation serves as the global node for all four deployment sites. The workstation is equipped with an AMD Ryzen Threadripper PRO 5975WX 32-core processor and two NVIDIA RTX A6000 GPUs. Unlike the edge nodes, which prioritize continuous low-latency inference under constrained resources, the global node provides the computational capacity required for centralized data management, statistical analysis, and scene-specific anomaly-model training.

All four edge nodes communicate with the same global node through the available network infrastructure. This configuration reflects the intended YUTELA deployment model in which inference resources are distributed across physical locations while computationally intensive model adaptation is consolidated on a shared global server.

\subsection{Operational Workflow and Communication}
\label{subsec:operational_workflow}

During deployment, live CCTV streams are continuously ingested by the corresponding Jetson Thor edge node. Camera streams from the same location are batched and distributed across the available AI pipelines. Real-time perception and anomaly inference are performed locally, with each camera maintaining its own scene-specific anomaly model. In parallel, selected anomaly features, contextual metadata, and pseudo-labeled training samples are associated with their originating location and camera ID and transferred to the global node for statistical analysis and scene-specific model adaptation. Once sufficient recent training data have been accumulated for a camera, the global node performs periodic retraining and returns the updated weights to the corresponding edge node.

Continuous raw video transmission to the global node or cloud is therefore not required during normal operation. Communication is primarily limited to selected features, training data, statistical information, model synchronization, and updated model weights. This design reduces network dependence for real-time inference while enabling centralized model adaptation across multiple deployment sites.

\section{Experimental Evaluation}
\label{sec:experiments}

We evaluate YUTELA through three complementary experiments that progress from controlled system benchmarking to long-term real-world operation. First, we benchmark the edge node using prerecorded surveillance video~\cite{dukemtmc} under different scene densities and camera concurrency levels. This experiment quantifies the computational performance of the embedded implementation and compares it with the workstation-based Ancilia system~\cite{ancilia,yao2025lab}. Second, we deploy the complete YUTELA framework at four real-world sites and operate the system continuously for a 170-hour, approximately one week, observation period. This experiment evaluates end-to-end runtime performance and the behavior of the centralized adaptation workflow under realistic workloads. Third, we evaluate the end-to-end processing latency of anomaly alerts triggered under real-world deployment conditions.

Together, these experiments address three system-level questions: 1) whether an embedded edge node can provide sufficient multi-camera processing capability to replace the workstation class hardware, 2) whether the complete distributed system can operate reliably across real-world deployments while supporting camera-specific model updates, and 3) whether the deployed system can deliver anomaly alerts with practical end-to-end latency under different real-world conditions.


\subsection{Edge-Node System Benchmark}
\label{subsec:edge_benchmark}

\begin{figure*}[t]
    \centering
    \includegraphics[width=0.85\textwidth]{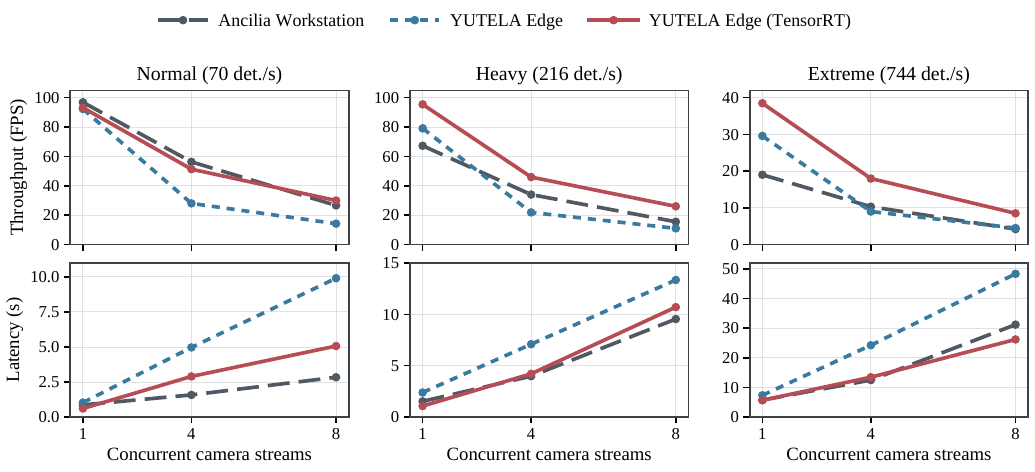}
    \caption{Edge-node throughput (top) and end-to-end pipeline latency (bottom) as the number of concurrent camera streams increases, under normal (70 detections/s), heavy (216 detections/s), and extreme (744 detections/s) crowd density. The Ancilia workstation (Threadripper Pro 3975WX, three A6000 GPUs) is compared with YUTELA on a single Jetson Thor, with and without TensorRT.}
    \label{fig:edge_scalability}
\end{figure*}

\subsubsection{Experimental Setup}

The first experiment isolates the edge processing node and evaluates its computational efficiency under controlled video workloads. We compare three system configurations: the original Ancilia implementation~\cite{ancilia} running on a workstation-class setup with a Threadripper Pro 3975WX CPU and three A6000 GPUs, YUTELA running on the NVIDIA Jetson Thor, and YUTELA running on the same Jetson Thor with TensorRT-optimized perception models. The second configuration includes the redesigned pipeline described in Section~\ref{subsec:scene_adaptation}, including camera redistribution, shared memory, reduced inter-stage data transfer, and stage integration, with the models executed in their original framework. The third configuration adds TensorRT acceleration to this pipeline, so the difference between the two YUTELA configurations isolates the effect of TensorRT. Because the Jetson Thor has substantially less compute than the workstation, the comparison with Ancilia reflects the combined effect of the hardware change and the software redesign.

To characterize the effect of scene complexity, we use prerecorded surveillance video representing three workload levels: normal, heavy, and extreme crowd density. The workloads contain approximately 70, 216, and 744 detections per second, respectively. Each workload is evaluated using 1, 4, and 8 concurrent camera streams. This controlled configuration allows the effect of camera concurrency and scene density to be evaluated independently of variations in network conditions or deployment duration.

For each configuration, we measure processing throughput and end-to-end pipeline latency. The experiment therefore evaluates both the computational scalability of the embedded implementation and the benefit provided by TensorRT optimization.

\begin{table*}[t]
  \centering
  \caption{Multi-site deployment, runtime performance, and triggered continual-training statistics.}
  \label{tab:multisite_runtime}

  \small
  \setlength{\tabcolsep}{3.5pt}
  \renewcommand{\arraystretch}{1.15}

  \begin{tabular}{@{}c|ccc|ccc|cccc@{}}
    \toprule
    \toprule

    \multirow{2}{*}{\textbf{Site}} &
    \multicolumn{3}{c|}{\textbf{Deployment}} &
    \multicolumn{3}{c|}{\textbf{Runtime Performance}} &
    \multicolumn{4}{c}{\textbf{Triggered Continual Training}} \\

    \cmidrule(lr){2-4}
    \cmidrule(lr){5-7}
    \cmidrule(lr){8-11}

    &
    \textbf{Cams.} &
    \textbf{Env.} &
    \makecell{\textbf{Input}\\\textbf{FPS}} &
    \makecell{\textbf{Peak}\\\textbf{Flow}} &
    \makecell{\textbf{Min Proc. FPS}\\\textbf{/Cam.}} &
    \makecell{\textbf{Max Lat. (s)}\\\textbf{/Cam.}} &
    \makecell{\textbf{Total}\\\textbf{Triggers}} &
    \makecell{\textbf{Triggers}\\\textbf{/Cam.}} &
    \makecell{\textbf{Inter-Trigger}\\\textbf{Interval (h)}} &
    \makecell{\textbf{Train Time}\\\textbf{/Event (min)}} \\

    \midrule

    Site A &
    8 & Urban Campus & 30 &
    29 & 10.96 & 16.71 &
    9 & 0–3 & 55.0 ± 20.4 & 4.2 ± 2.4 \\

    Site B &
    7 & Dining Complex & 30 &
    63 & 2.98 & 37.38 &
    59 & 6–10 & 19.4 ± 4.6 & 7.7 ± 4.4 \\

    Site C &
    8 & Public Marketplace & 15 &
    77 & 2.78 & 28.03 &
    20 & 1-4 & 42.2 ± 6.3 & 6.2 ± 3.4 \\

    Site D &
    8 & Center & 20 &
    30 & 1.92 & 15.98 &
    13 & 0-5 & 41.0 ± 12.8 & 4.6 ± 3.1 \\

    \bottomrule
    \bottomrule
  \end{tabular}
\end{table*}

\subsubsection{Edge-Node Performance}

Fig.~\ref{fig:edge_scalability} plots the throughput and latency of the three system configurations as the number of camera streams grows, with one panel per crowd density. Three trends are visible. First, directly moving the pipeline to the embedded platform (blue dotted line) causes a clear drop in throughput and a steep rise in latency as streams are added, even though its single-stream performance is similar to or better than that of the workstation. Second, TensorRT optimization (red solid line) lifts throughput above the workstation (gray dashed line) in seven of the nine settings, and the gap widens as scene density increases. Third, latency grows with the number of streams for all configurations. With four or eight streams under normal and heavy density, the optimized edge node still has higher latency than the workstation, so the gain from the edge platform is mainly in throughput rather than in response time.

For the normal density workload with eight camera streams, TensorRT increases throughput from 14.17 FPS to 29.94 FPS compared with the unoptimized YUTELA implementation, an improvement of approximately 111\%. At the same time, latency decreases from 9.91~s to 5.07~s. The optimized edge implementation also slightly exceeds the 26.60 FPS achieved by the Ancilia workstation under the same workload, although its latency remains higher than the workstation's 2.84~s.

The improvement is also clear under higher scene density. With eight streams under the heavy-density workload, TensorRT increases throughput from 10.97 FPS to 25.97 FPS, compared with 15.48 FPS on the original Ancilia workstation, while latency (10.70~s) remains close to the workstation's 9.54~s. Under the extreme-density workload, the optimized edge implementation reaches 8.48 FPS, compared with 4.47 FPS without TensorRT and 4.18 FPS on the Ancilia workstation. Latency is also reduced from 48.33~s to 26.19~s compared with the unoptimized edge implementation, which is the only eight-stream setting in which the edge node is faster than the workstation (31.19~s). More generally, the advantage of the optimized edge pipeline grows with scene density: with eight streams, its throughput is 1.13, 1.68, and 2.03 times that of the workstation under normal, heavy, and extreme density, respectively. This is likely because per-person stages such as pose estimation and anomaly inference dominate the workload in crowded scenes and benefit most from TensorRT acceleration.

Overall, the results show that simply moving Ancilia from a workstation to an embedded device is not enough to maintain its original performance. With inference optimization, however, YUTELA can support multi-camera processing on the edge while achieving performance comparable to, and in some cases higher than, the workstation-based system. The results also show that both the number of camera streams and scene density have a strong effect on throughput and latency, which motivates the long-term real-world evaluation presented next.


\begin{figure*}[t]
    \centering
    \subfloat[Site A -- Urban College]{%
        \includegraphics[width=0.45\textwidth]{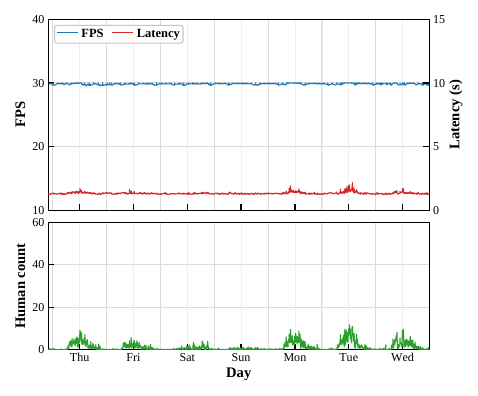}%
        \label{fig:site_a}}
    \hfill
    \subfloat[Site B -- Dining and Nightlife Complex]{%
        \includegraphics[width=0.45\textwidth]{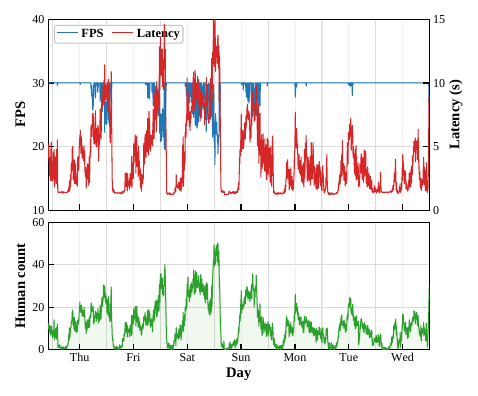}%
        \label{fig:site_b}}

    \vspace{-4pt}
    \subfloat[Site C -- Public Marketplace]{%
        \includegraphics[width=0.45\textwidth]{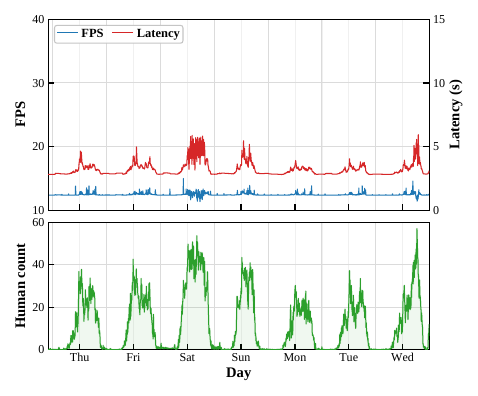}%
        \label{fig:site_c}}
    \hfill
    \subfloat[Site D -- Community Arts Center]{%
        \includegraphics[width=0.45\textwidth]{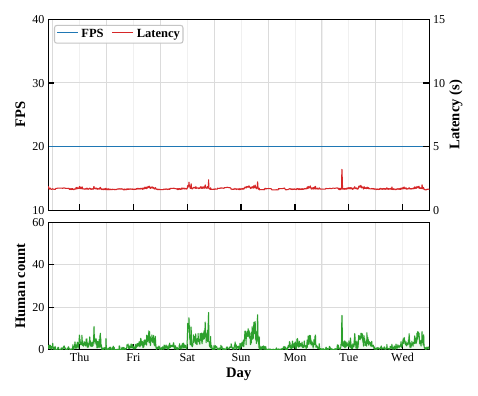}
        \label{fig:site_d}}
    \caption{Runtime behavior of YUTELA during the continuous 170-hour deployments at the four sites. In each panel, the upper plot shows processing throughput (FPS) and pipeline latency, and the lower plot shows the number of people detected across the site's cameras.}
    \label{fig:deployment_sites}
\end{figure*}

\subsection{Real-World Multi-Site Deployment}
\label{subsec:realworld_deployment}

\subsubsection{Deployment Setup}

The second experiment evaluates the complete YUTELA system during continuous real-world operation. YUTELA is deployed at four local sites with different camera layouts, scene characteristics, activity levels, and operating conditions. Unlike the controlled edge-node benchmark, this experiment covers the full system workflow, including camera ingestion, edge inference, anomaly processing, and centralized periodic training.

The four sites are described in Section~\ref{subsec:deployment_sites}: an urban college (Site A), a dining and nightlife complex (Site B), a public marketplace (Site C), and a community arts center (Site D). Together they use 31 cameras, seven or eight per site, with input frame rates from 15 to 30 FPS. All sites are evaluated over the same continuous 170-hour observation period.

During deployment, we record both runtime system performance and adaptation activity. Runtime measurements include the minimum processing rate observed per camera and the maximum pipeline latency. For adaptation, we record the total number of training triggers, the number of triggers for each camera, the interval between successive training events, and the execution time of each training process.

\subsubsection{Long-Term Runtime Performance}

Table~\ref{tab:multisite_runtime} summarizes the deployment settings, runtime performance, and periodic-training activity observed across the four sites. The results show that YUTELA remains operational under different real-world workloads, while throughput and latency vary with both scene activity and camera input settings.

Site A represents the campus deployment with eight 30-FPS camera streams. The system reaches a minimum processing rate of 10.96 FPS per camera and a maximum observed latency of 16.71~s. Although pedestrian activity on campus is generally continuous, most cameras are directed toward road crossings, sidewalks, and outward-facing areas rather than dense indoor spaces. As a result, the number of people detected within the camera views remains relatively limited, reducing the processing load compared with more crowded sites.

Site B provides a more demanding workload. The downtown restaurant contains seven 30-FPS camera streams and experiences higher and more frequent pedestrian activity. The minimum processing rate decreases to 2.98 FPS per camera, while the maximum latency reaches 37.38~s. As shown in Fig.~\ref{fig:site_b}, the stronger variation in pedestrian flow is closely reflected in the runtime behavior, making this deployment a useful example of YUTELA operating under sustained higher load.

Sites C and D also show that camera frame rate affects system latency in addition to scene activity. Site C, the food-market deployment, has pedestrian flow comparable to Site B in several periods, but its cameras provide 15-FPS input streams. The TensorRT inference pipeline uses a fixed 30-frame input batch, requiring the system to wait longer to form a complete batch from a 15-FPS stream. This introduces approximately one additional second of data collection before inference and contributes to a different latency pattern from the 30-FPS restaurant deployment, even when the observed traffic level is similar.

A similar effect is observed at Site D, the art-center deployment, where the cameras operate at 20 FPS and pedestrian activity is generally lower. Its runtime behavior falls between the higher-frame-rate campus deployment and the lower-frame-rate food-market deployment. Together, Sites A--D show that real-world performance depends not only on the number of cameras and the number of detected people, but also on the input frame rate and the time required to prepare inference batches.


Fig.~\ref{fig:deployment_sites} shows the long-term runtime behavior of YUTELA at all four sites over the same week. The lower plot in each panel traces the daily rhythm of each location, and the upper plot shows how the edge pipeline responds to it.

The peak pedestrian-flow measurements further show the differences between the four workloads. Site C reaches the highest peak flow of 77, followed by Site B at 63, while Sites A and D reach peaks of 29 and 30, respectively. These results show that camera count alone does not fully describe the processing load of a deployed surveillance system. Scene activity, input frame rate, and the number of detected people can all change the amount of work required by the edge.

The same deployments are also used to evaluate YUTELA's periodic scene-specific training process. Across the four sites, the system generates 101 training triggers during the 170-hour observation period, although the number and timing of these events differ considerably between sites.

Site A generates nine training events, with an average inter-trigger interval of $55.0 \pm 20.4$~h. Site B generates 59 events and has the shortest average interval of $19.4 \pm 4.6$~h. Sites C and D generate 20 and 13 events, with average intervals of $42.2 \pm 6.3$~h and $41.0 \pm 12.8$~h, respectively. These differences reflect the data-volume-based periodic adaptation strategy introduced in our previous work~\cite{yao2026offline}: training is triggered by the amount of usable scene-specific data collected by each camera rather than by a fixed update schedule.

Despite the variation in trigger frequency, individual training events remain relatively short. The average training time ranges from $4.2 \pm 2.4$~min at Site A to $7.7 \pm 4.4$~min at Site B. Because training is performed on the centralized global server rather than on the Jetson Thor edge nodes, model updates do not occupy the edge GPU used for continuous video inference.

Overall, the long-term deployment shows two important characteristics of YUTELA. First, runtime performance changes with the actual scene workload and camera configuration rather than camera count alone. Second, the amount of scene-specific training data collected varies across deployments, causing different cameras and sites to require model updates at different times. This supports the use of a data-driven training trigger instead of applying the same fixed retraining schedule to every deployment.

\subsection{End-to-End Alert Processing Latency}
\label{subsec:alert_latency}

In addition to frame-processing throughput, we evaluate the end-to-end latency of real world anomaly alert generation. Following the PCP evaluation procedure introduced in~\cite{yao2024evaluating}, the measurement covers the complete processing path from camera input to the final anomaly alert, rather than the execution time of an individual model. Both object level and behavior level anomaly processing are evaluated.

To conduct the experiment under representative operating conditions, we visited all four deployment sites during their typical busy periods, selected based on the traffic patterns observed during the long-term deployment experiment. At each site, object-anomaly and behavior-anomaly events were triggered separately. Each scenario was repeated 15 times, and the resulting end-to-end latency was recorded to calculate the mean, standard deviation, and observed range.

\begin{table}[t]
  \centering
  \caption{Physical-Cyber-Physcial alert-trigger latency for object and behavior anomalies with eight concurrent surveillance cameras under different traffic flows. The Ancilia baseline is reported from \cite{yao2024evaluating}.}
  \label{tab:pcp_latency_comparison}

  \footnotesize
  \setlength{\tabcolsep}{2.4pt}
  \renewcommand{\arraystretch}{1.10}

  \begin{tabular}{@{}c|c|c|c|cc@{}}
    \toprule
    \toprule

    \textbf{System} &
    \textbf{Location} &
    \makecell{\textbf{Traffic}\\\textbf{Flow}} &
    \textbf{Anomaly} &
    \makecell{\textbf{Mean $\pm$ Std.}\\\textbf{(s)}} &
    \makecell{\textbf{Range}\\\textbf{(s)}} \\

    \midrule

    \multirow{2}{*}{Ancilia \cite{yao2024evaluating}} &
    \multirow{2}{*}{Location B} &
    \multirow{2}{*}{0--10} &
    Object &
    $8.92 \pm 0.71$ &
    7.46--10.61 \\

    & & &
    Behavior &
    $11.62 \pm 0.62$ &
    10.75--13.30 \\

    \midrule

    \multirow{8}{*}{YUTELA} &
    \multirow{2}{*}{Site A} &
    \multirow{2}{*}{0--10} &
    Object & $6.82 \pm 0.57$ & 5.70--7.69 \\

    & & &
    Behavior & $8.71 \pm 0.63$ & 7.87--10.27 \\

    \cmidrule(lr){2-6}

    &
    \multirow{2}{*}{Site B} &
    \multirow{2}{*}{20--30} &
    Object & $11.33 \pm 0.69$ & 9.98--12.39 \\

    & & &
    Behavior & $15.41 \pm 0.74$ & 13.96--16.54 \\

    \cmidrule(lr){2-6}

    &
    \multirow{2}{*}{Site C} &
    \multirow{2}{*}{20--30} &
    Object & $8.45 \pm 0.80$ & 6.88--9.67 \\

    & & &
    Behavior & $11.61 \pm 0.86$ & 9.92--12.93 \\

    \cmidrule(lr){2-6}

    &
    \multirow{2}{*}{Site D} &
    \multirow{2}{*}{0--10} &
    Object & $7.64 \pm 1.02$ & 5.64--9.20 \\

    & & &
    Behavior & $9.78 \pm 0.92$ & 7.98--11.19 \\

    \bottomrule
    \bottomrule
  \end{tabular}
\end{table}

Table~\ref{tab:pcp_latency_comparison} summarizes the results together with the previous measurement reported in~\cite{yao2024evaluating}. In that experiment, Ancilia achieved an object anomaly latency of $8.92 \pm 0.71$~s and a behavior anomaly latency of $11.62 \pm 0.62$~s using eight concurrent cameras at Location B.

For YUTELA, the measured latency varies across deployment sites and traffic conditions. Under lower traffic flow, such as Sites A and D, object anomaly latency remains between approximately 6.8 and 7.6~s on average, while behavior anomaly latency remains below 10~s. Under the higher traffic conditions observed at Sites B and C, both object and behavior processing require more time, reflecting the additional workload introduced by larger numbers of detected people and downstream features. Among the four deployments, Site A achieves the lowest average object-anomaly latency at $6.82 \pm 0.57$~s, while Site B shows the highest average latency at $11.33 \pm 0.69$~s for object anomalies and $15.41 \pm 0.74$~s for behavior anomalies. Sites C and D fall between these two cases. These results are consistent with the long term runtime evaluation, where scene activity and camera configuration were shown to directly affect the processing workload of the edge pipeline.

The measurements should not be interpreted as a direct one to one comparison between sites or with the previous Ancilia experiment, since the deployments use different cameras, environments, input frame rates, and traffic conditions. Instead, the experiment provides a practical view of the alert latency that can be expected when YUTELA operates in real world environments. The results also suggest that the redesigned edge pipeline can maintain relatively low alert latency under lighter traffic, while latency increases as scene activity and processing demand become higher.


\subsection{Lessons Learned and Limitations}
\label{subsec:experiment_discussion}

The three experiments evaluate YUTELA from edge efficiency to complete real-world operation, and together they suggest several lessons for deploying video anomaly detection on existing city camera infrastructure.

\textbf{Edge hardware trades latency for throughput:}
The controlled benchmark shows that the Jetson Thor can replace a workstation with three A6000 GPUs for multi-camera processing, but only after TensorRT optimization. The optimized edge node exceeds the workstation's throughput in seven of nine settings, and its advantage grows with crowd density. However, with four or eight streams under normal and heavy density, its latency remains higher than the workstation's. For applications that require the shortest possible response time at low load, a workstation remains faster. For deployments across many sites, the edge node offers comparable or higher throughput on a much smaller platform.

\textbf{Camera count is a poor proxy for workload:}
The long-term deployment shows that throughput and latency follow pedestrian activity, detected-object flow, and input frame rate, not the number of cameras. Site B, with seven cameras, was the most demanding site, while Site A, with eight cameras at the same frame rate, maintained the highest minimum processing rate. Input frame rate also interacts with batching: at the 15-FPS marketplace site, filling a fixed 30-frame batch adds about one second before processing begins. Edge resources should therefore be planned from expected scene activity and camera configuration rather than from camera count alone.

\textbf{Data-driven retraining triggers adapt to each scene:}
Because retraining is triggered by the amount of collected pose data rather than by a fixed schedule, its frequency followed scene activity, from a mean interval of 19.4~h at Site B to 55.0~h at Site A. A fixed schedule would either retrain quiet cameras with little new data or update busy cameras too rarely. The centralized training load was small: all 101 events together occupied about 6.6\% of one GPU over the observation period, and the linear projection suggests that a single global node can support tens of sites.

\textbf{Limitations:}
This study evaluates system efficiency, multi-site operation, and the adaptation workflow, but not whether the updated models detect anomalies more accurately than the initial ones. Measuring this would require manually verified anomaly labels for every camera scene, which is outside the scope of this work. Because retraining relies on pseudo-normal data, anomalous activity that is not detected may be included in the training data and gradually absorbed into the model's notion of normal behavior. User feedback on false and missed alerts, discussed in Section~\ref{sec:conclusion}, is one way to reduce this risk. The capacity projection assumes that training jobs run sequentially and does not include data-upload time or bursts of simultaneous triggers. Finally, the alert-latency measurements were taken at different sites, with different cameras and traffic, from the earlier Ancilia measurement, and should not be read as a direct comparison.


\section{Conclusion}
\label{sec:conclusion}

This work presented YUTELA, an evolution of the Ancilia intelligent video surveillance system for multi-site AIoT deployment. YUTELA moves on-site processing from workstation-class servers to NVIDIA Jetson Thor edge devices and redesigns the multi-camera pipeline through camera redistribution, shared memory, reduced inter-stage data transfer, stage integration, and TensorRT acceleration. In controlled benchmarks with eight streams, TensorRT raised throughput from 14.17 to 29.94 FPS, and the optimized edge node exceeded the throughput of the workstation-based system in seven of nine workload settings, by up to $2.03\times$ at extreme density, although latency remained higher under lighter loads.

Beyond edge inference, YUTELA maintains a separate anomaly model for each camera and retrains it on a centralized GPU server when enough recent pseudo-normal data have been collected, without placing the training workload on the embedded platform. A continuous 170-hour deployment at four public sites with 31 existing CCTV cameras showed that runtime load follows scene activity and input frame rate rather than camera count, that data-driven retraining frequency varied by about three times across sites, and that the 101 resulting training events occupied only about 6.6\% of one GPU. Mean end-to-end alert latency ranged from 6.8 to 15.4~s depending on site traffic and anomaly type. Overall, YUTELA provides a practical and low-cost framework for combining efficient edge inference with camera-specific model adaptation in long-term, multi-site surveillance deployments.

Future work will measure the effect of adaptation on detection accuracy using manually verified evaluation data for each camera scene, and will extend YUTELA toward human-in-the-loop learning. In particular, user feedback from the alerting interface can be used to identify and label false-positive events, following the feedback-based learning strategy explored in \cite{yao2025alfred}. Such feedback could improve pseudo-label quality and guide future model updates. Additional work will investigate larger deployment scales, scheduling of simultaneous retraining jobs on the global node, improved data selection strategies, and further reductions in communication and training overhead.


\section*{Acknowledgment}

This work was supported by the U.S. National Science Foundation (NSF) under Award No.~2527312, ``CIVIC-FA Track B: Strengthening Community Safety and Well-being with Ethical, AI-Assisted Video Solutions.'' Any opinions, findings, and conclusions or recommendations expressed in this material are those of the authors and do not necessarily reflect the views of the National Science Foundation.


\bibliographystyle{IEEEtran}
\bibliography{reference}





\end{document}